\pdfoutput=1

\documentclass[11pt]{article}

\usepackage{acl}

\usepackage{times}
\usepackage{latexsym}
\usepackage{amsmath}
\usepackage{graphicx}
\usepackage{booktabs}
\usepackage{tabularray}
\usepackage{array}
\usepackage{ragged2e}
\usepackage{multirow}
\usepackage{amsfonts}
\usepackage{colortbl}      
\usepackage[ruled,vlined]{algorithm2e}
\usepackage{setspace}
\usepackage{hyperref}
\usepackage{caption} 
\usepackage{subcaption} 
\usepackage[english]{babel}
\usepackage{wasysym}
\usepackage{longtable}
\usepackage{array}

\definecolor{commentcolor}{RGB}{110,154,155}   
\newcommand{\PyComment}[1]{{\small\ttfamily\textcolor{commentcolor}{\# #1}}}  
\newcommand{\PyCode}[1]{{\small\ttfamily\textcolor{black}{#1}}} 

\usepackage[T1]{fontenc}

\usepackage[utf8]{inputenc}

\usepackage{microtype}
 
\title{LoraRetriever: Input-Aware LoRA Retrieval and Composition for Mixed Tasks in the Wild}

\author{
Ziyu Zhao\textsuperscript{1},
Leilei Gan\textsuperscript{1},
Guoyin Wang\textsuperscript{2},
Wangchunshu Zhou\textsuperscript{3},\\
Hongxia Yang\textsuperscript{2},
Kun Kuang\textsuperscript{1},
Fei Wu\textsuperscript{1} \\
\texttt{benzhao.styx@gmail.com}, \\
\texttt{\{leileigan, kunkuang, wufei\}@zju.edu.cn}, \\
\texttt{guoyinwang.duke@gmail.com, hx.yang@bytedance.com, chunshu@aiwaves.cn}, \\
\textsuperscript{1}Zhejiang University,
\textsuperscript{2}ByteDance Inc.,
\textsuperscript{3}AIWaves Inc. \\
}

\begin{document}
\maketitle
\begin{abstract}

Low-Rank Adaptation (LoRA) provides an effective yet efficient solution for fine-tuning large language models (LLMs). The modular and plug-and-play nature of LoRA enables the integration of diverse domain-specific LoRAs to enhance the capabilities of LLMs. Previous research on exploiting multiple LoRAs either focuses on specific isolated downstream tasks or fixes the selection of LoRAs during training. However, in real-world scenarios, LLMs receive diverse prompts covering different tasks, and the pool of candidate LoRAs is often dynamically updated. To bridge this gap, we propose LoraRetriever, a retrieve-then-compose framework that adaptively retrieves and composes multiple LoRAs according to the input prompts. LoraRetriever contains three main components: firstly, identifying and retrieving LoRAs relevant to the given input; secondly, formulating strategies for effectively integrating the retrieved LoRAs; and thirdly, developing efficient batch inference to accommodate heterogeneous requests. Experimental results indicate that LoraRetriever consistently outperforms the baselines, highlighting its practical effectiveness and versatility.
\end{abstract}

\section{Introduction}

Recently, large language models (LLMs) such as ChatGPT~\cite{liu2023gpt} and Llama~\cite{touvron2023llama} have shown notable successes in a range of fields~\cite{hadi2023survey,wang2023survey}. Nevertheless,  due to the prohibitively high computation costs for fine-tuning LLMs on specific domains, there is a growing shift towards Parameter-Efficient Fine-Tuning (PEFT)~\cite{liu2022p, hu2023llm, hu2021lora}, which only updates a small fraction of the model's parameters or integrates new trainable parameters that augment the model's capabilities. Within this sphere, Low-Rank Adaptation (LoRA)~\cite{hu2021lora} stands out for its remarkable effectiveness, modularity, and plug-and-play capabilities. As AI communities like \href{https://huggingface.co/}{Huggingface} and \href{https://modelscope.cn/home/}{ModelScope} witness an influx of LoRA parameters tailored for diverse tasks and domains, there is an increasing emphasis on employing various LoRA experts to provide a comprehensive service~\cite{sheng2023s}.
 
\begin{figure}[t]
    \centering
    \includegraphics[width=.85\linewidth]{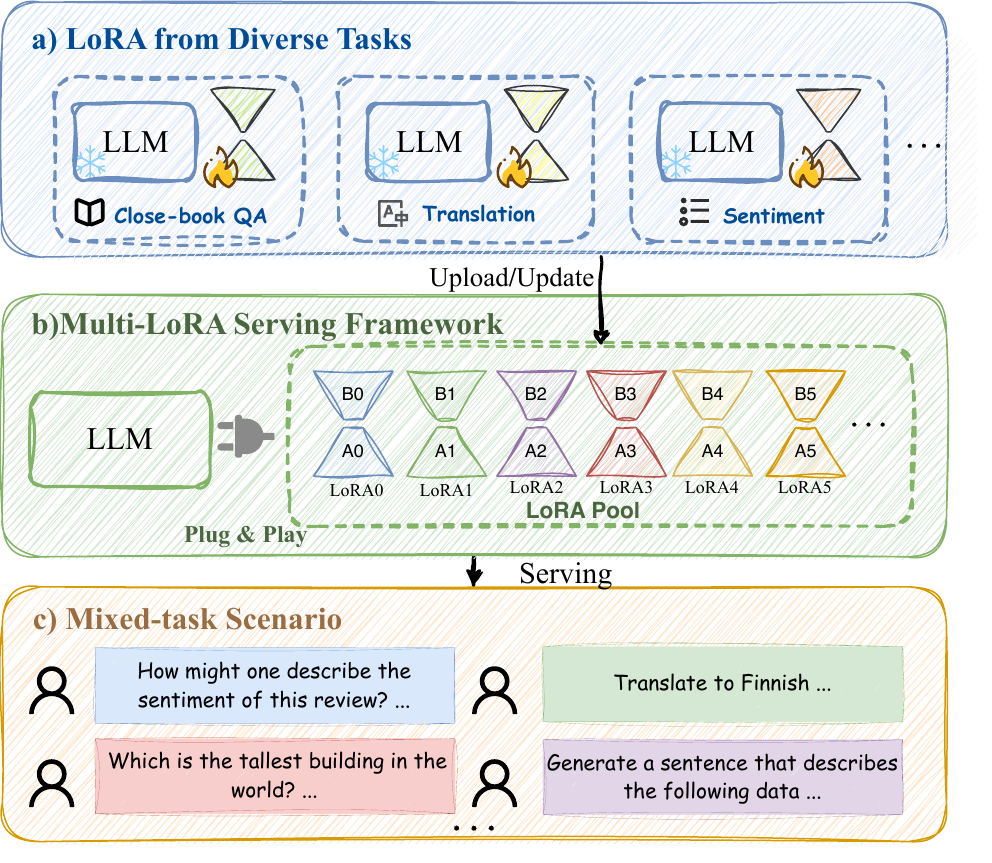}
    \vspace{-0.1in}
    \caption{Illustration of serving multiple LoRAs within a dynamically updated LoRA pool for mixed-task scenarios. a) LoRAs from various domains and tasks aimed at enhancing specific capabilities of the LLM can be uploaded to or updated to the LoRA pool. b) The multi-LoRA serving framework aims to leverage the plug-and-play nature of LoRAs to offer comprehensive services. c) The downstream tasks, presented in a mixed-task form, require personalized expert routing.}
    \vspace{-0.25in}
    \label{fig:mix_task}
\end{figure}

Recent research has explored the integration of the mixture of expert (MoE;~\citet{jacobs1991adaptive,jordan1994hierarchical}) with LoRAs~\cite{wang2022adamix,liu2023moelora,zadouri2023pushing,muqeeth2023soft,anonymous2024mole}.
However, these methods lock in the selection of LoRAs while training, lacking the ability to dynamically update and scale in scenarios where the LoRA pool may consistently expand.
LoRAhub~\cite{huang2023lorahub} and AdapterSoup~\cite{chronopoulou2023adaptersoup} explore composing LoRAs for specific downstream tasks.
However, these two methods offer a one-size-fits-all solution for downstream tasks, overlooking the heterogeneous nature of diverse real-world requests. 

To bridge these gaps, our paper explores the "mixed-task scenario" as exemplified by platforms like ChatGPT~\cite{liu2023gpt} and Gemini~\cite{team2023gemini}, wherein the nature of user requests encompasses diverse prompts covering different tasks. 
While LLMs present a unified solution for a broad spectrum of tasks, their performance can still falter in certain specialized areas\cite{liu2023moelora,yang2023fingpt}. This is where the integration of LoRAs becomes crucial.
As shown in Fig.\ref{fig:mix_task}, our vision encompasses a multi-LoRA serving framework capable of dynamic enhancement, continuously improving its functionality as new LoRA modules are added and updated. Through the plug-and-play capabilities of LoRA, the framework can provide personalized services for heterogeneous downstream requests.



In this paper, we introduce LoraRetriever, a retrieve-then-compose framework designed to exploit the plug-and-play nature of LoRA in mixed-task scenarios. Our framework consists of three key components: (1) \textbf{Input-aware LoRA Retrieval:} The first step of our framework is aligning user inputs with the corresponding LoRAs through sentence embeddings and is further refined by an instruction fine-tuning~~\cite{su2022one,asai2022task} for effective LoRA retrieval. 
Through the retriever, we achieve a more flexible LoRA routing mechanism, whose training stage is disentangled from the training and inference of the LLM.
(2) \textbf{LoRA Composition:}
Our framework next employs two strategies for compositing the retrieved LoRAs in the first step. The \textit{Fusion of LoRAs} averages multiple LoRAs' parameters and constructs a singular comprehensive model for each input. The \textit{Mixture of LoRAs} activates multiple LoRAs simultaneously and then averages the output of each submodule of the LoRAs. Composing top-$k$ LoRAs increases the recall rate for the correct LoRA and improves the generalization of unseen tasks by integrating the LoRAs of similar tasks. (3) \textbf{Batch Inference:} Most previous work on the input-adaptive inference of LLMs does not support batch inference~\citep{zhou2020bert, chronopoulou2023adaptersoup}. To tackle the challenge of heterogeneous batched requests, we construct a unique LoRA mapping matrix for batch samples. This allows for tailored inferences through efficient matrix multiplication, ensuring each request activates its corresponding LoRAs while maintaining batch processing efficiency.

To assess the performance of LoraRetriever, we established a mixed-task evaluation benchmark comprising 48 LoRAs spanning a variety of natural language understanding and generation tasks. The experimental results underline the effectiveness of the proposed methods in serving both in-domain and out-of-domain downstream requests. Furthermore, the retrieval routing method exhibits a robust generalization capability: although the retriever is trained on just 40\% of the tasks, it effectively retrieves the corresponding LoRAs for unseen tasks.


\section{Related Work}
\textbf{Mixture of Experts.} 
The Mixture of Experts (MoE) method combines various specialized sub-modules, guided by a gating network to tailor responses to different input types~\cite{jacobs1991adaptive,jordan1994hierarchical,shen2023mixture,riquelme2021scaling,dou2023loramoe}. 
Some work~\cite{wang2022adamix, zadouri2023pushing,zhu2023sira, liu2023moelora,dou2023loramoe} focuses on using the MoE method for PEFT to achieve more effective and efficient model fine-tuning.
Other work~\cite{anonymous2024mole,muqeeth2023soft} focuses on using MoE to coordinate existing LoRA experts without specifically training the experts' parameters.
These methods require training additional parameters for gating and are limited to a fixed number of LoRAs, making them unsuitable for complex and dynamic scenarios. Our method can be seen as gating through a retriever, hence achieving flexibility and generalization on unseen experts.

\textbf{Adapter Merging.}
In addition to model ensembling through the MoE, there is an increasing focus on aggregating adapters from different domains through the method of Adapter Merging. 
AdapterSoup~\cite{chronopoulou2023adaptersoup} aggregates different adapters in the parameter space, allowing large language models to adapt to new domains without additional training. 
LoRAhub~\cite{huang2023lorahub} employs random sampling of LoRA parameters from various domains and tasks, followed by black-box optimization to learn the weights of different LoRA parameters without involving model gradient backpropagation. 
These methods offer a one-size-fits-all solution for downstream tasks, which cannot be applied in the mixed-task scenario for providing personalized service.

\section{Preliminaries}
This section begins with a concise introduction to the Low-Rank Adaptation, followed by a detailed formalization of the mixed-task scenario.

\subsection{Low-Rank Adaptation}
Directly fine-tuning large language models with all parameters is computationally intensive and is not feasible in low-resource scenarios. Based on the idea that only a small number of low-rank parameters need to be fine-tuned for sufficient performance in new domains, \citet{hu2021lora} proposed the Low-Rank Adaptation, where the LoRA module can be combined with the pre-trained parameters in parallel for efficient inference. 

Specifically, given pre-trained weights $W_0 \in \mathbb{R}^{d\times d}$ of a sub-module of LLM, the LoRA adds an extra trainable weight matrix as $W_0 + \Delta W = W_0 + BA$, where $\Delta W$ can be decomposed into two smaller matrices $B\in \mathbb{R}^{d\times r}$ and $A \in \mathbb{R}^{r \times d}$, where $r$ stands for the rank of $\Delta W$. The forward pass can be modified as follows:
\begin{equation}
    x^{\prime} = W_0 x+\Delta Wx = W_0 x + BAx,
\end{equation}
where $x \in \mathbb{R}^d$ is the input and the $x^{\prime} \in \mathbb{R}^d$ denote the output.

\subsection{Problem Formulation}
In this part, we give a formal definition of the mixed-task scenario. 
Given an original LLM $L$, we have a set of $k$ LoRAs, $\Phi=\{\phi_1, \phi_2, \cdots, \phi_k\}$, on the shelf, where each LoRA $\phi_i$ is trained on its corresponding task $T_i$.
The mixed task inputs can be formulated as $T_{mix} = \{x, \forall x \in T_1 \lor T_2 \cdots \lor T_k \}$, where $\lor$ stands for the logical disjunction operator. 

Under the mixed-task scenario, given an input $x \in T_{mix}$ without its task tag, the serving process can be written as:
\begin{equation}
\label{eq:framework}
    y = F(g(\Phi, x), x, \theta),
\end{equation}
where $\theta$ denotes the original parameters of LLM, $g(\Phi, x)$ represents the input-aware LoRA retrieval process and returns a set of retrieved LoRAs $\Phi_i$.
$F(\Phi_i, x_i, \theta)$ depicts the LoRA composition process that integrates the retrieved LoRAs as a plug-in to the original LLM.


\begin{figure*}
    \centering
    \includegraphics[width=.78\linewidth]{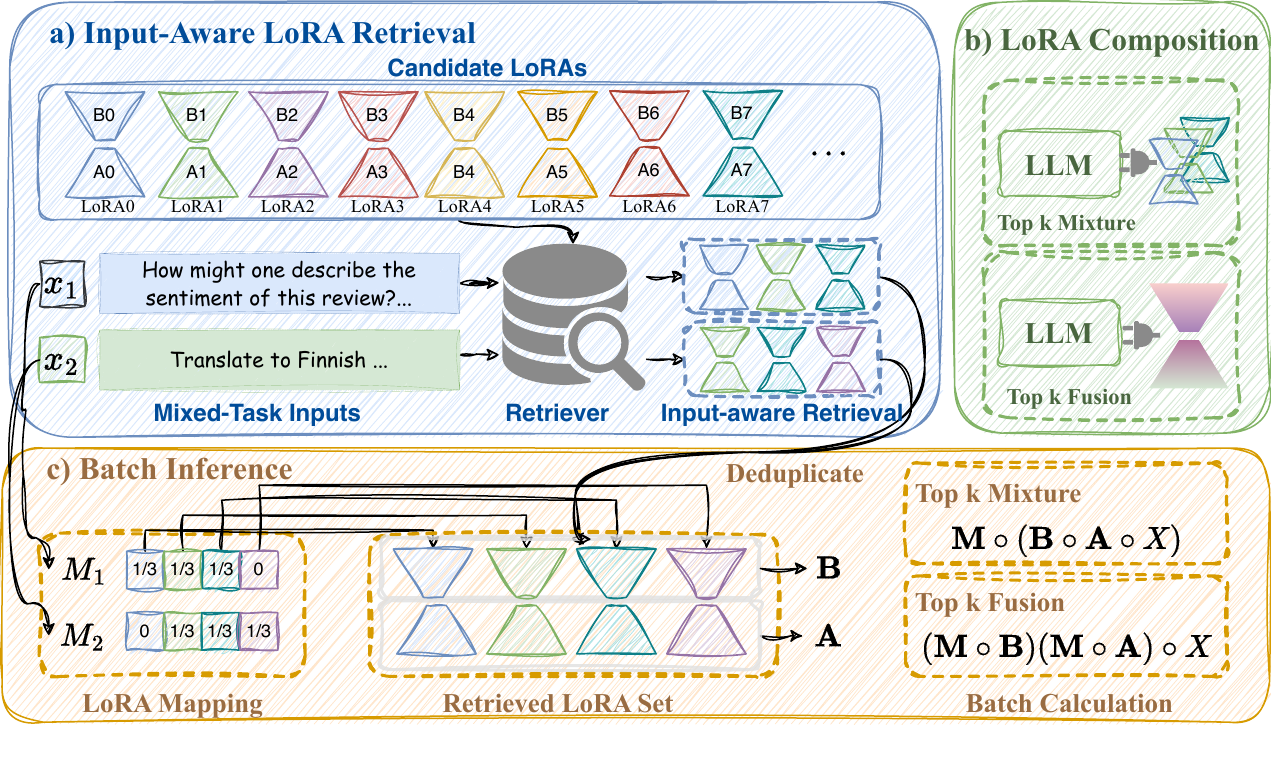}
    \vspace{-0.2in}
    \caption{The LoraRetriever Framework. This framework, equipped with a pool of candidate LoRAs from various domains/tasks, is designed to offer personalized services tailored to the input provided. It begins by executing an input-aware LoRA retrieval process aimed at identifying LoRAs corresponding to tasks analogous to the input (\S\ref{sec:lora_retrieval}). Subsequently, it employs a specialized LoRA composition mechanism to efficiently utilize the retrieved LoRAs (\S\ref{sec:lora_composition}). By constructing a LoRA mapping matrix for batch inputs, the framework facilitates effective batch inference (\S\ref{sec:batch}).
 }
    \label{fig:framework}
    \vspace{-0.2in}
\end{figure*}

\section{LoraRetriever Framework}
In this section, we describe the LoraRetriever framework as shown in Fig.\ref{fig:framework} for serving multi-LoRAs in mixed-task scenarios. This framework contains three major components: the input-aware LoRA retrieval module (\S\ref{sec:lora_retrieval}), the LoRA composition module (\S\ref{sec:lora_composition}), and the batch inference strategy (\S\ref{sec:batch}).

\subsection{Input-Aware LoRA Retrieval}
\label{sec:lora_retrieval}
Our goal is to construct a LoraRetriever tailored to effectively retrieve the corresponding LoRAs for each input in scenarios where LoRAs are dynamically updated. However, existing approaches fall short of accurately identifying LoRAs under such conditions. MoE-based methods ~\cite{anonymous2024mole,muqeeth2023soft} struggle to generalize when new LoRAs are introduced due to the fixed selection of LoRAs established during router training. Retrieval methods like sentence embedding~\cite{reimers2019sentence,ni2021sentence} or task embedding~\cite{achille2019task2vec, zhou-etal-2022-efficiently} fail to map both samples and LoRA into a shared embedding space, limiting their effectiveness in input-aware LoRA retrieval.


To achieve this goal, we propose to train a retriever via instruction fine-tuning ~\cite{su2022one,asai2022task}, namely LoraRetriever, which can retrieve suitable LoRAs from a massive LoRA pool for a given input sample. The fundamental concept behind LoraRetriever comprises two main steps: (i) First, to embed different task-specific LoRAs into embedding space for facilitating retrieval, we posit that each LoRA can be represented by some data points, which can be obtained by randomly choosing a dozen samples from the training dataset. Then we average their instruction embeddings to represent the embedding of each LoRA. (ii) 
To improve generalization for unseen LoRAs in LoRA retrieving, we train the retriever through instruction fine-tuning~\cite{su2022one,wei2021finetuned} on a subset of all tasks. Training on a small subset of tasks is designed to simulate scenarios involving the integration of new LoRAs, thereby underscoring our method's generalization abilities via instruction fine-tuning. These two strategies enable the effective use of limited data distributions for input-aware retrieval and can be generalized to unseen LoRAs.

Formally, with a sentence-embedding model $E$, input sequence $x$, and the instruction $I$ for embedding purposes, the instructed embedding can be formulated as $E(I \oplus x)$, where $\oplus$ denotes the concatenation operation. In order to allow the embedding to capture the similarity between different tasks, the instruction is expressed as: \textit{\colorbox{red!10}{"Represent the sentence for similar task retrieval"}}. Each LoRA module is embedded with $m$ randomly selected domain-specific samples, expressed as $E(\phi)=\frac{1}{m}\sum^m_{i=1}E(I \oplus x_{i\phi})$. This embedding method integrates both sample-wise and LoRA-module-wise embeddings, facilitating the calculation of similarity between an individual sample and a LoRA module. For measuring the similarity between LoRA module $\phi$ and the input sequence $x$, following ~\cite{ni2021sentence}, we leverage the cosine similarity between the LoraRetriever embeddings: $s(x, \phi, I) = \cos(E(I \oplus x), E(\phi))$.

To improve LoRA retrieval by the retriever and broaden its generalization to unseen LoRAs, we train the embedding model $E$ through instruction fine-tuning on a small subset of tasks. To prevent the need to access new samples, we use previously employed samples for embedding LoRAs as our training data.
Consider $t$ distinct training tasks, represented as $\mathcal{T}_{train} = \{T_1, \cdots, T_t\}$. Following \citet{ni2021sentence}, the training dataset $\mathcal{D}$ comprises paired samples ${(x_i, x_i^+)}$, where each $x_i$ is a sample from a task $T_i \in \mathcal{T}_{train}$, and a positive sample $x_i^+$ is randomly selected from the same task $T_i$. To complement each positive pair, we randomly select $p$ negative pairs ${(x_i, x_{ij}^-)}_{j=1}^p$, ensuring that $x_{ij}^-$ is sourced from tasks outside of $T_i$, thereby $x_{ij}^- \notin T_i$.
The training process is achieved through a contrastive loss~\cite{karpukhin2020dense, izacard2021unsupervised,ni2021sentence} defined as follows:
\begin{equation*}
    \mathcal{L} = \frac{e^{s(x_i,x_i^+, I)/\gamma}}{e^{s(x_i,x_i^+, I)/\gamma} + \sum_{j=1}^p e^{s(x_i,x_{ij}^-, I)/\gamma}},
\end{equation*}
where $\gamma$ is the softmax temperature.

During the LoRA retrieval phase, the top-$k$ LoRAs are retrieved according to their similarity to the input $x$. This process can be formulated as follows:
\begin{equation*}
    g(x_i, \Phi) :=  \Phi_i = \mathrm{TopK} \{s(\phi_j,x_i,I) , \phi_j \in \Phi \}.
\end{equation*}

\subsection{LoRA Composition}
\label{sec:lora_composition}
After retrieving the top-k LoRAs $\Phi_i$ for an input $x_i$, we proceed to integrate these LoRAs into the LLM with parameter $\theta$. This integration is achieved by applying two different LoRA composition strategies: the \textit{Mixture of LoRAs} and the \textit{Fusion of LoRAs}.

\subsubsection{Mixture of LoRAs}
The mixture of LoRAs strategy involves the aggregation of the outputs of each submodule within the assembled LoRAs. Let us denote $\mathbf{A} = \{A_1, A_2, \ldots, A_n\}$ and $\mathbf{B} = \{B_1, B_2, \ldots, B_n\}$ as the sets representing submodules within $n$ LoRAs. For an input $x_i$, the output derived from the mixture of LoRAs can be expressed as $x_i^{\prime} = \frac{1}{n} \sum_{j=1}^n B_j A_j x_i$,
where $x_i^{\prime}$ denotes the output. This process signifies the integration of each LoRA module's output, effectively blending their contributions to form a unified output.

\subsubsection{Fusion of LoRAs}
In contrast to the Mixture method, which combines the output of different LoRAs, fusing the parameters of these LoRAs presents an alternative composition strategy.

Let the parameters of each LoRA $\phi_i$ be denoted by $\Theta_i$. The parameter of the fused LoRA is then represented as $\Theta_{\text{fusion}} = \frac{1}{k} \sum_{j=1}^k \Theta_j$. This formulation allows the fused parameter to function akin to a single LoRA.

\begin{table*}
\centering

\resizebox{.8\linewidth}{!}{  
\begin{tabular}{l c c c c c c c c c c c c c} 
\toprule
\multirow{2}{*}{Task} &
\multirow{2}{*}{\begin{tabular}[c]{@{}c@{}}Perfect\\Selection\end{tabular}} &
\multicolumn{2}{c}{Selection} & 
\multicolumn{2}{c}{Fusion} & 
\multicolumn{2}{c}{Mixture} & 
\multirow{2}{*}{\begin{tabular}[c]{@{}c@{}}MoE\\Top1\end{tabular}} &
\multirow{2}{*}{\begin{tabular}[c]{@{}c@{}}MoE\\Top3\end{tabular}} &
\multirow{2}{*}{\begin{tabular}[c]{@{}c@{}}MoE\\Soft\end{tabular}} &
\multirow{2}{*}{\begin{tabular}[c]{@{}c@{}}SME-\\AR\end{tabular}} &
\multirow{2}{*}{\begin{tabular}[c]{@{}c@{}}Adapter\\Soup\end{tabular}} &
\multirow{2}{*}{\begin{tabular}[c]{@{}c@{}}LoRA\\Hub\end{tabular}}\\
\cline{3-8}
& & IID & OOD & IID & OOD & IID & OOD & & & & \\
\midrule
\multicolumn{14}{c}{\textit{w/ Llama2-7b}} \\\hline
$\text{Struct to Text}_{Rouge-1}$ & \cellcolor[gray]{0.8} 64.0 & \textbf{61.3} & 50.1 & 49.4 & 45.9 & 55.9  & \underline{50.4} & 45.6 & 46.8 & 47.9   & 48.0      & 4.5 & 35.6 \\
$\text{Struct to Text}_{Rouge-2}$  & \cellcolor[gray]{0.8} 39.6 & \textbf{37.0} & \underline{26.6} & 25.7 & 23.5 & 30.0  & 26.4 & 21.9 & 22.9 & 23.8    & 24.2      & 1.1 & 17.7 \\
$\text{Struct to Text}_{Rouge-l}$  & \cellcolor[gray]{0.8} 57.0 & \textbf{54.5} & 43.9 & 43.6 & 40.3 & 49.5  & \underline{44.0} & 39.8 & 40.7 & 41.7    & 42.4      & 4.5 & 31.6 \\
$\text{Translation}_{BLEU}$ & \cellcolor[gray]{0.8} 13.1 & \textbf{12.8} & 12.0 & 12.2 & \underline{12.3} & \textbf{12.8}  & 12.2 & 9.5 & 10.5 & 10.7    & 11.0      & 1.4 & 8.5 \\
\midrule
COMMONSENSE & \cellcolor[gray]{0.8} 62.5 & 55.5 & 46.0 & 51.0 & 48.0 & \textbf{61.5}  & \underline{50.0} & 54.5 & 52.0 & 51.5    & 50.0      & 46.0     & 17.5 \\
SENTIMENT & \cellcolor[gray]{0.8} 90.0 &\textbf{89.5} & 89.0 & 79.0 & 78.5 & \textbf{89.5} & \underline{90.5} & 70.0 & 75.0 & 74.5    & 74.0      & 73.5     & 0.5 \\
READING Comp. & \cellcolor[gray]{0.8} 67.3 & \textbf{51.7} & 40.3 & 47.3 & 45.0 & 51.3  & \underline{47.3} & 48.7 & 47.7 & 48.7    & 45.7      & 40.7     & 2.7 \\
CLOSE-BOOK QA & \cellcolor[gray]{0.8} 45.0 & 40.0 & 43.0 & 41.0 & 37.5 & \textbf{45.0}  & \underline{48.5} & 40.5 & 38.5 & 40.0    & 32.0      & 31.5     & 1.0 \\
COREFERENCE & \cellcolor[gray]{0.8} 52.0 & 50.0 & 46.0 & 47.0 & \underline{53.0} & \textbf{63.0}  & 49.0 & 61.0 & 59.0 & 57.0    & 58.0      & 43.0     & 1.0 \\
READ. COOMP. W/ COM & \cellcolor[gray]{0.8} 69.0 & \textbf{69.0} & 30.0 & 35.0 & 19.0 & 46.0  & \underline{40.0} & 31.0 & 29.0 & 29.0    & 23.0      & 14.0     & 3.0 \\
PARAPHRASE & \cellcolor[gray]{0.8} 65.5 & \textbf{58.0} & \underline{45.5} & 45.5 & 44.0 & 56.5  & \underline{45.5} & 42.0 & 38.5 & 36.0    & 34.5      & 46.5     & 1.0 \\
NLI & \cellcolor[gray]{0.8} 72.3 & \textbf{70.0} & 60.6 & 51.4 & 53.8 & 67.9  & \underline{64.3} & 50.3 & 49.6 & 48.3    & 50.8      & 62.4     & 10.5 \\
\midrule
\multicolumn{14}{c}{\textit{w/ Llama2-13b}} \\\hline
$\text{Struct to Text}_{Rouge-1}$ & \cellcolor[gray]{0.8} 65.4 & \textbf{62.6} & 49.4 & 52.7 & 49.7 & 57.7  & \underline{52.1} & 46.8    & 47.0    & 48.5    & 48.3 & 7.1 & 39.3     \\
$\text{Struct to Text}_{Rouge-2}$  & \cellcolor[gray]{0.8} 40.8 & \textbf{38.2} & 25.8 & 29.2 & 26.8 & 32.6  & \underline{28.1} & 24.5    & 25.1    & 25.7    & 25.2 & 2.5 & 20.7     \\
$\text{Struct to Text}_{Rouge-l}$  & \cellcolor[gray]{0.8} 58.7 & \textbf{56.0} & 42.9 & 45.9 & 43.2 & 50.8  & \underline{45.4} & 41.1    & 41.9    & 42.7    & 42.2 & 6.4 & 34.6     \\
$\text{Translation}_{BLEU}$ & \cellcolor[gray]{0.8} 12.9 & 12.9 & 12.7 & \textbf{14.6} & \underline{14.1} & \textbf{14.6}  & \underline{14.1} & 11.8    & 12.4    & 11.9    & 12.4 & 0.8 & 10.2     \\
\midrule
COMMONSENSE & \cellcolor[gray]{0.8} 69.5 & 59.0 & 47.5 & 61.0 & 56.0 & 64.0  & \underline{60.5} & 65.0    & \textbf{66.0}    & 64.0    & 61.0 & 17.5     & 34.0     \\
SENTIMENT & \cellcolor[gray]{0.8} 90.0 & 90.5 & 91.0 & 87.0 & 83.5 & \textbf{91.5}  & \underline{91.5} & 90.0    & 89.5    & 90.0    & 89.0 & 79.5     & 11.0     \\
READING Comp. & \cellcolor[gray]{0.8} 76.0 & \textbf{60.3} & 48.0 & 56.7 & 49.3 & \textbf{60.3}  & \underline{51.3} & 53.7    & 53.3    & 52.3    & 51.3 & 48.7     & 3.3 \\
CLOSE-BOOK QA & \cellcolor[gray]{0.8} 64.0 & 60.0 & 53.0 & 62.0 & 58.0 & \textbf{63.0}  & \underline{61.0} & 59.5    & 57.5    & 58.5    & 57.5 & 34.5     & 6.5 \\
COREFERENCE & \cellcolor[gray]{0.8} 74.0 & 75.0 & \underline{65.0} & 55.0 & 59.0 & \textbf{76.0}  & 64.0 & 61.0    & 62.0    & 56.0    & 57.0 & 55.0     & 10.0     \\
READ. COOMP. W/ COM & \cellcolor[gray]{0.8} 82.0 & \textbf{80.0} & 33.0 & 57.0 & 49.0 & 78.0  & \underline{58.0} & 51.0    & 48.0    & 49.0    & 49.0 & 13.0     & 14.0     \\
PARAPHRASE & \cellcolor[gray]{0.8} 77.5 & 68.0 & 52.5 & 55.5 & 45.5 & \textbf{71.0}  & \underline{55.5} & 50.0    & 52.5    & 47.5    & 52.0 & 64.0     & 2.5 \\
NLI & \cellcolor[gray]{0.8} 82.4 & \textbf{78.9} & 70.2 & 69.8 & 66.4 & 78.1  & \underline{75.7} & 67.7    & 71.0    & 67.4    & 66.6 & 67.5     & 14.9     \\
\bottomrule
\end{tabular}
}
\caption{We report the average performance of each task cluster. The full results of each task are shown in Appendix.\ref{sec:full_results}. 
"IID" signifies that LoraRetriever can access any LoRA for every test sample, encompassing the LoRA specific to the sample's task. "OOD" indicates that for each test sample, we mask the LoRA associated with its specific task during the retrieval phase. Consequently, no sample can access its ideal LoRA, allowing us to assess the LoraRetriever's cross-task generalization capability. The performance of perfectly selected corresponding LoRA for each sample is colored in gray. We have bolded the best performance of each task and underlined the best performance in the "OOD" setting.}
\vspace{-0.15in}
\label{tab:main_sum}
\end{table*}

\begin{table}[htb]
\centering
\resizebox{.85\linewidth}{!}{  
\begin{tabular}{lcccc}
\toprule
\textbf{Method} & \textbf{Top 1} & \textbf{Top 3} & \textbf{Top 5} & \textbf{Top 8} \\ 
\midrule
all-mpnet-base-v2 & 58.40 & 78.26 & 84.77 & 90.24 \\
all-MiniLM-L6-v2 & 51.73 & 73.11 & 80.54 & 87.18 \\
msmarco-distilbert-cos-v5 & 45.84 & 66.01 & 75.14 & 82.67 \\
gtr-t5-xl & 53.19 & 69.72 & 77.41 & 83.59 \\
\midrule
LoraRetriever \textsuperscript{0\%} & 60.80 & 79.29 & 85.57 & 91.58 \\
 \cellcolor[gray]{0.8}
 LoraRetriever \textsuperscript{40\%} & \cellcolor[gray]{0.8} 63.16 & \cellcolor[gray]{0.8} 89.09 & \cellcolor[gray]{0.8} 95.45 & \cellcolor[gray]{0.8} 98.97 \\
LoraRetriever \textsuperscript{100\%} & \textbf{74.08} & \textbf{97.37} & \textbf{99.15} & \textbf{99.82} \\
\bottomrule
\end{tabular}
}
\vspace{-0.1in}
\caption{Comparison of Sentence Embedding Techniques in LoRA Retrieval: The notation LoraRetriever\textsuperscript{k\%} signifies that the model underwent supplementary training on $k$ percent of the tasks. The performance of the selected retriever model in the evaluation phase is highlighted in gray.}
\vspace{-0.2in}
\label{tab:retrival_acc}
\end{table}

\subsection{Batch Inference of Multiple LoRAs}
\label{sec:batch}
Implementing batch inference in the presence of multiple LoRAs and diverse composition diagrams poses a significant technical challenge. To address this, we introduce a unique approach for batch inference. Our method involves processing a batch of samples denoted as $X \in \mathbb{R}^{b \times l \times d}$, where $b$, $l$, and $d$ denote the batch size, sequence length, and sample dimensionality, respectively. 
For each input $x_i$ and its retrieved LoRAs $\Phi_i$ within the same batch, we aggregate these LoRAs into a collective set denoted by $\Phi_\mathcal{B}$. To ensure the uniqueness of $\Phi_\mathcal{B}$, we eliminate duplicates, mindful of the possibility that retrieved LoRAs may overlap across different samples. The resulting set $\Phi_\mathcal{B}$ comprises $p$ unique LoRAs, where $p \leq bk$. For every sample $x_i$, a $p$ dimension mapping vector $M_i$ is generated, which specifies the indices of its corresponding LoRAs within $\Phi_\mathcal{B}$.

The LoRA mapping vectors are combined into a matrix $\mathbf{M} \in \mathbb{R}^{b\times p}$. The parameters of a submodule in LoRA can be denoted as $A$ and $B$, and are concatenated within the batched LoRAs $\Phi_{\mathcal{B}}$ to obtain $\mathbf{A} \in \mathbb{R}^{p\times r\times d}$ and $\mathbf{B} \in \mathbb{R}^{p\times d\times r}$.
The batch inference process of the mixture of LoRAs can be formulated as follows:
\begin{equation}
    X^{\prime} = \mathbf{M}\circ (\mathbf{B} \circ\mathbf{A} \circ X),
\end{equation}
where we denote the batched output of a layer of multiple LoRA as $X^{\prime}\in \mathbb{R}^{b \times l \times d}$ and extend the symbol $\circ$ to denote potential broadcasting as~\citet{wen2023batched}. The batch inference process of LoRA fusion can be formulated as
\begin{equation}
    X^{\prime} = (\mathbf{M} \circ \mathbf{B})(\mathbf{M} \circ\mathbf{A}) \circ X.
\end{equation}
These strategies can be simply implemented by the einsum operation, and the PyTorch-style pseudocode is shown in Appendix.\ref{sec:pseudocode}.

\section{Experiments}
This section outlines the evaluation framework for assessing different approaches in mixed-task scenarios. Furthermore, a comprehensive analysis of the proposed LoraRetriever framework is presented.

\subsection{Evaluation Framework}
\paragraph{Base Model \& LoRA Configuration.} To test various methods in the mixed-task scenarios, we leverage Llama-2-\{7b,13b\}~\cite{touvron2023llama} as the base models and train a range of LoRAs for a spectrum of tasks. We select a portion of the Flan-v2 datasets~\cite{wei2021finetuned} to train 48 LoRAs for a spectrum of tasks covering Natural Language Understanding (NLU) and Natural Language Generation (NLG). Following the categorization by \citet{wei2021finetuned}, these tasks can be grouped into 10 distinct task clusters. We train each LoRA according to the Alpaca~\cite{alpaca} format and rank $r$, and the scaling hyperparameter $\alpha$ are set to 6 and 12, respectively. 
The details of the LoRAs training can be found in the Appendix. \ref{sec:eval_dataset}.

\paragraph{Mixed Task Evaluation Dataset.} For constructing the mixed-task dataset, we randomly chose 50 samples from the test set for each task used in training 48 LoRAs, subsequently mixing and shuffling these samples to form a unified dataset with 6000 data entries. Further details about these datasets are available in the Appendix.\ref{sec:eval_dataset}. 

\paragraph{Baseline Methods.} We compared our method with the following baselines: 
(1) \textbf{Mixture of Experts}
~\cite{zhu2023sira,zadouri2023pushing,liu2023moelora,wang2022adamix,anonymous2024mole}. 
(2) \textbf{SMEAR} ~\cite{muqeeth2023soft};
(3) \textbf{AdapterSoup} ~\cite{chronopoulou2023adaptersoup};
(4) \textbf{LoRAhub} ~\cite{huang2023lorahub}.
Specifically, we implement three variants of MoE.
A detailed description of the baseline models can be found in Appendix.\ref{sec:baseline}, with their implementations presented in Appendix.\ref{sec:moe_train}.

\paragraph{Implementation of LoraRetriever.} To train the LoraRetriever, we continue to perform instruction fine-tuning based on Instructor-xl ~\cite{su2022one}. The training data consisted of only 40\% of the tasks used to train task-specific LoRAs, with each task represented by 20 samples randomly selected from its respective LoRA training set.
In this process, we categorized samples from the same LoRA as positive examples, while those from different LoRAs were considered negative examples. Additionally, the three distinct strategies for LoRA composition are: (1) \textbf{Selection}, which involves choosing the highest-ranked (top-1) retrieved LoRA and applying it in a singular LoRA manner, and can be viewed as a variant for Mixture and Fusion methods; (2) \textbf{Mixture}, which averaging the outputs of each submodule from the top-$k$ retrieved LoRAs; and (3) \textbf{Fusion}, a method that averages the parameters of the top-$k$ retrieved LoRAs. Throughout our experiments, $k=3$ is established as the default setting.

\begin{table}
\centering
\resizebox{.85\linewidth}{!}{  
\begin{tabular}{lccc}
\toprule
Methods & 0\% & 40\% ($\Delta$\%) & 100\% ($\Delta$\%) \\ \midrule
Selection & 57.99 & 62.42 (+7.64\%) & 64.01 (+2.55\%) \\
Fusion & 51.50 & 51.50 (+0.00\%) & 52.27 (+1.49\%) \\
Mixture & 62.24 & 63.54 (+2.09\%) & 64.19 (+1.02\%) \\
\bottomrule
\end{tabular}
}
\vspace{-0.1in}
\caption{Average Performance of LoraRetriever on NLU Tasks Across Different LoRA Composition Strategies and Task Training Percentages.}
\vspace{-0.25in}
\label{table:methods_improvement}
\end{table}

\paragraph{Metrics.} Following \citet{wei2021finetuned}, we assess the performance on the "Struct to Text" task using Rouge-\{1, 2, L\} and on the "Translation" tasks using BLEU. Additionally, for the NLU tasks, we evaluate the exact match accuracy of each method.

\subsection{Main Results}
The main results of the mixed-task evaluation are shown in Tab.\ref{tab:main_sum}. We present the mean performance across each task cluster and additionally evaluate the LoraRetriever's effectiveness in an out-of-domain (OOD) setting.
In the OOD configuration, we mask the corresponding LoRA for each sample, thereby inhibiting LoraRetriever from retrieving the ideal LoRA for each sample. In this way, we can assess the cross-task generalization capability of LoraRetriever. 
From the results, we have the following observations: (1) The proposed framework, LoRARetriever, which performs input-aware LoRA retrieval and composition, markedly surpasses other baselines focusing on specific downstream tasks. (2) Among them, the performance of Mixture and Selection is similar in IID scenarios, while Fusion's performance is weaker compared to the other two methods. The reasons are as follows: (i) In the IID setting, LoraRetriever can achieve strong top-1 selection, leading to similar results between Selection and the Mixture; (ii) As different tasks are inherently heterogeneous, it is inferior to directly average top-$k$ LoRA parameters in the Fusion. (3) In the OOD setting, the Mixture exceeds the performance of the Selection, and the performance of Fusion is similar to that of the Selection. The reasons can be as follows: (i) The selection cannot retrieve the associated LoRA for the input sample in the OOD setting, leading to a significant performance drop. (ii) The Mixture can fully leverage the capabilities of similar tasks to address OOD tasks, alleviating the performance drop. (4) The performance of the MoE and SMEAR methods is weaker than that of LoraRetriever. The limitation stems from the restricted capacity of these methods for adaptation and generalization to dynamically changing environments populated with diverse LoRAs, thereby diminishing their efficacy in mixed-task scenarios.
(5) In mixed-task scenarios, although AdapterSoup uniformly searches for appropriate LoRAs for downstream tasks, the retrieved LoRAs fall short in personalization for each request, hindering their effectiveness for each specific task. 
(6) LoRAhub proves to be entirely ineffective in the mix-task scenario. First, the fusion of LoRAhub depends on randomly selected LoRAs, which may not be relevant. Second, the presence of heterogeneous tasks introduces conflicting parameter optimization directions, resulting in the total breakdown of parameter fusion.

\subsection{Analysis}
\paragraph{Performance of Retriever.} 
We compare LoraRetriever with some popular off-the-shelf sentence embedding models in Huggingface and adopt the model nomenclature following \citet{wolf2020transformers}. 
To analyze the effect of the percentage of the tasks for training LoraRetriever, we trained three variants of LoraRetriever with different percentages.
Tab.\ref{tab:retrival_acc} shows the performance of different retrieval models for retrieving relevant LoRAs. It is shown that guiding sentence embedding models with specific prompts leads to a performance improvement in retrieval compared to common retrieval models. After instruction fine-tuning, the retriever significantly enhanced the ability to retrieve corresponding LoRA based on the input. Conducting instruction fine-tuning on 40\% of the tasks resulted in a 2.36\% increase in top-1 accuracy and a 9.80\% increase in top-3 accuracy. Training across all tasks achieved the largest improvement. To demonstrate the generalizability of the proposed framework when dealing with unseen LoRAs, we used a retriever trained on 40\% of the tasks in the main experiment to simulate the scenario of dynamic updates to the LoRA pool that might occur while providing services with LoraRetriever.

Tab.\ref{table:methods_improvement} shows the performance of LoraRetriever trained with different proportions of tasks for LoRA retrieval. It is observed that for the selection and mixture methods, training under 40\% of the tasks has already seen significant improvements. The best performance is achieved when trained under all tasks, but the improvement compared to 40\% is relatively small, which to some extent reflects the good generalization ability of instruction fine-tuning of LoraRetriever.

\begin{figure}
    \centering
    \includegraphics[width= .85\linewidth]{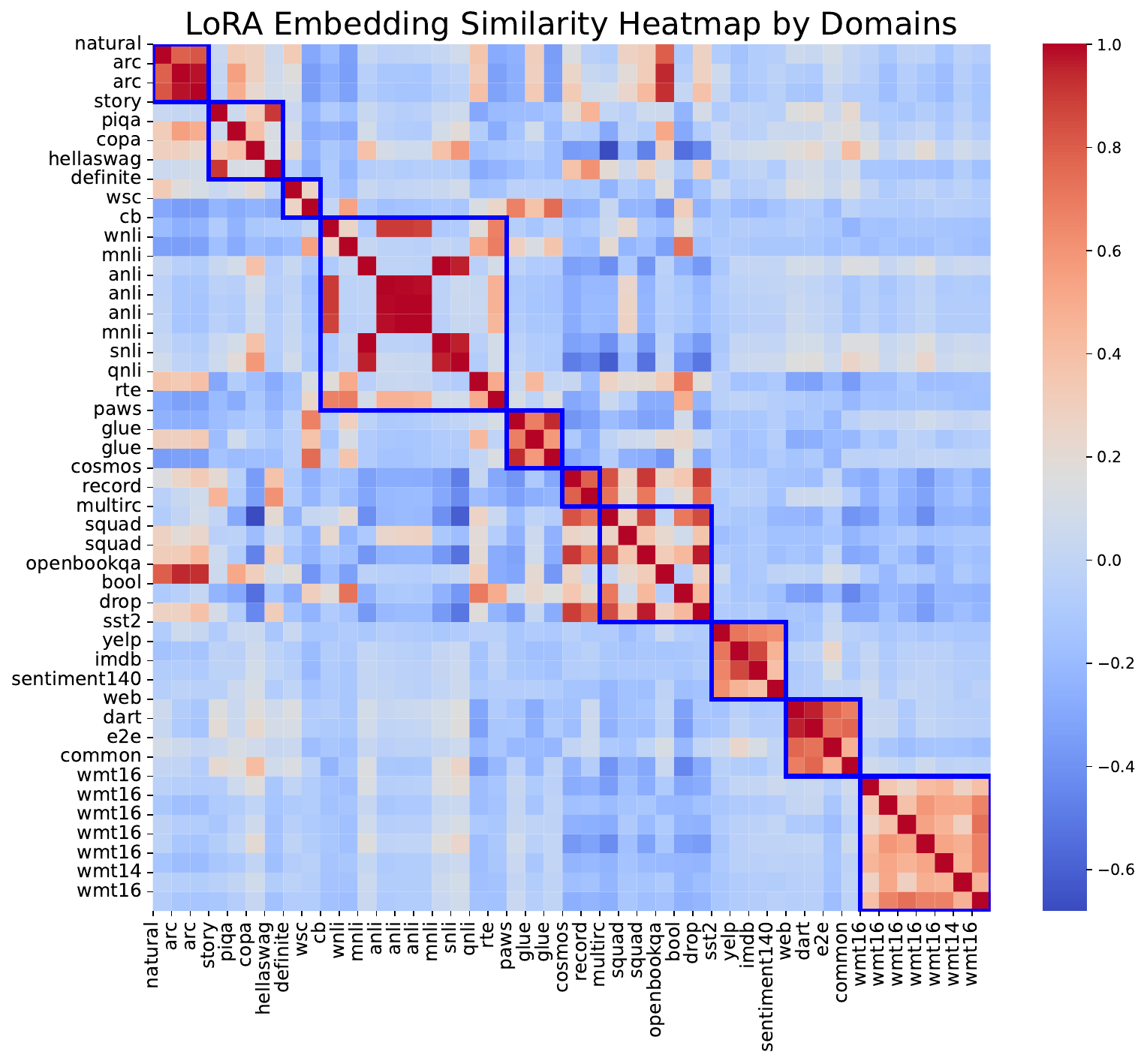}
    \vspace{-0.15in}
    \caption{LoRA embedding similarity heatmap. Tasks from the same domain are grouped in square brackets.}
    \vspace{-0.15in}
\label{fig:similarity_heatmap}
\end{figure}

Fig.\ref{fig:similarity_heatmap} illustrates the similarity between task embeddings for different tasks through a heatmap, where tasks from the same task cluster are grouped in square brackets. It is shown that task embeddings within the same domain are more similar, indicating that the LoraRetriever embeddings can serve as task embeddings to characterize the similarities between different tasks and be applied for LoRA retrieval.

\begin{figure}
    \centering
\includegraphics[width=.9\linewidth]{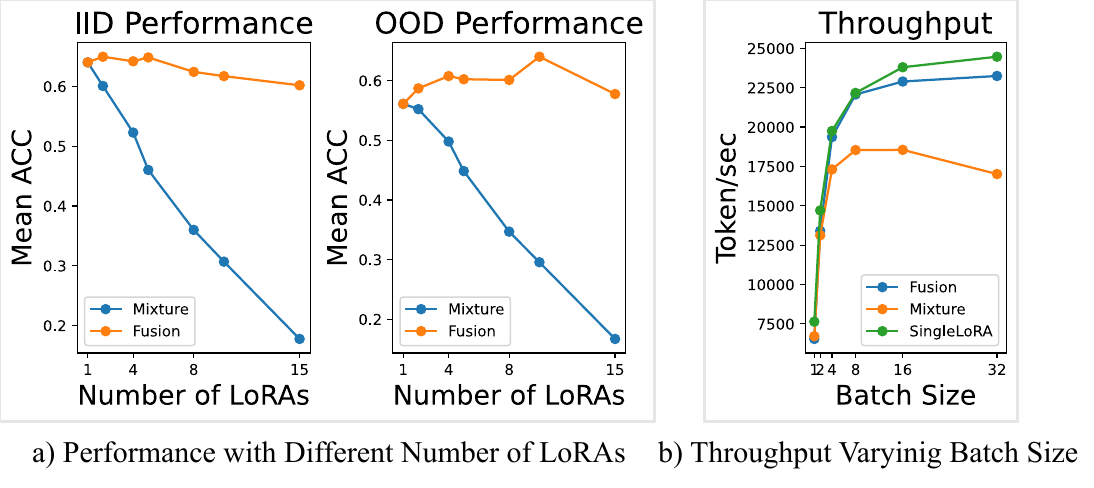}
    \caption{The left figure shows the performance of LoraRetriever varying the number of LoRAs. The right figure shows the performance of Throughput varying the batch size.}
    \vspace{-0.18in}
    \label{fig:loranum}
\end{figure}

\paragraph{Impact of the number of Retrieved LoRA.} Fig.\ref{fig:loranum} (a) illustrates the performance of the number of retrieved LoRAs on the mean accuracy of the NLU tasks. The results indicate that as the number of retrieved LoRAs increases, the performance of the Mixture initially improves slightly but then stabilizes. In contrast, the Fusion shows a continuous decline in performance with an increasing number of LoRAs, which once again demonstrates that under the conditions of heterogeneous tasks, the simple averaging of parameters can compromise the original capabilities of the LoRAs. In particular, in the OOD setting, the performance of the Mixture improves significantly as the number of LoRAs increases, illustrating that in the absence of an ideal LoRA choice for a request, leveraging the capabilities of multiple LoRAs of similar tasks can effectively achieve cross-task generalization.



\paragraph{Effectiveness of Batch Inference Strategy.}
To evaluate the efficiency of our proposed batch inference strategy, we compared the throughput of different batch sizes. The throughput is defined as the number of both input and output tokens per second across all requests in the mixed-task benchmark. We specifically compared the computational efficiency with that of a single LoRA. Our evaluation encompassed the entire evaluation dataset, and we limited the generation to the first produced token to mitigate discrepancies caused by varying generation lengths across different methods. These experiments were conducted on an NVIDIA A100 GPU (80GB) utilizing bfloat16 precision. As illustrated in Fig.\ref{fig:loranum} (b), our batch inference strategy markedly improves the throughput of the framework, with a slight throughput reduction compared to a single LoRA. Notably, the Fusion outperforms the mixture strategy in throughput efficiency, attributed to its parameter averaging approach that circumvents the need for parallel computation across multiple LoRAs.

\begin{figure}
    \centering
    \includegraphics[width=.85\linewidth]{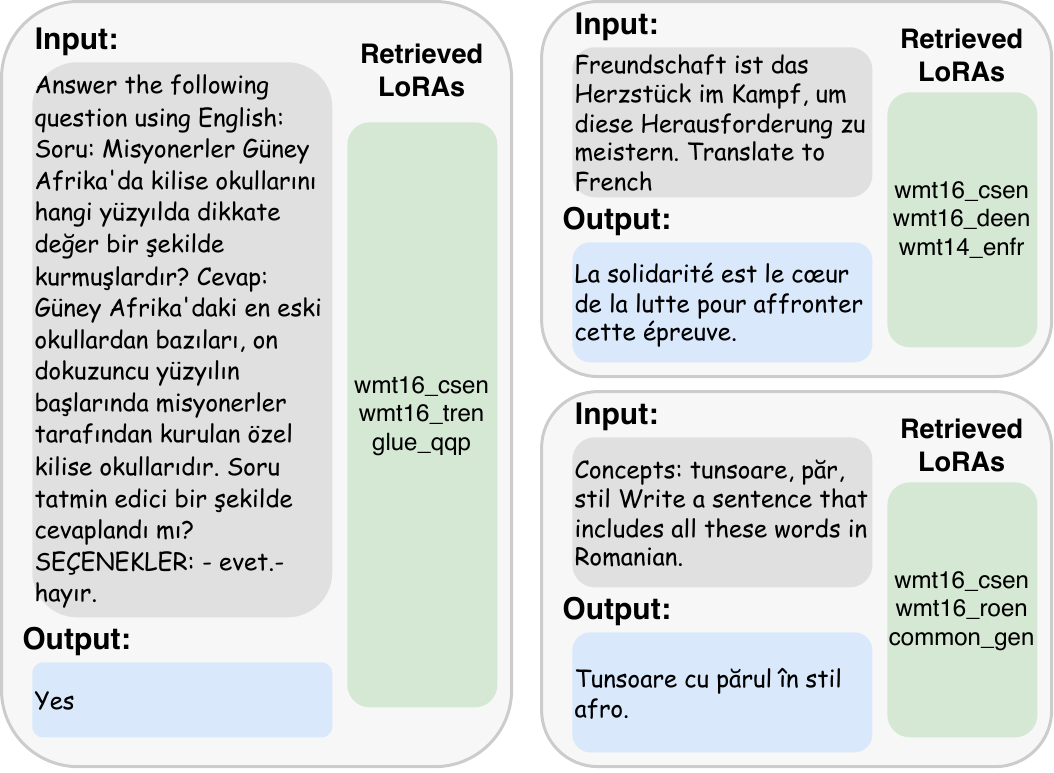}
    \vspace{-0.1in}
    \caption{Showcasing How the LoRARetrieval Framework Employs Multiple LoRAs for Cooperative Problem Solving.}
\label{fig:lora_ability_fusion}
    \vspace{-0.2in}
\end{figure}

\paragraph{Showcases.} We showcase the framework's ability to adeptly integrate multiple LoRAs for synergistic problem-solving, as evidenced in Fig.\ref{fig:lora_ability_fusion}. We manually craft three problems in Fig.\ref{fig:lora_ability_fusion}, which cannot retrieve any single LoRA to solve these problems directly, necessitating the cooperation of existing LoRAs.
Specifically, the first example requires LoraRetriever to integrate NLI and translation tasks' capabilities. The retrieved LoRA wmt16-tren is utilized for comprehending Turkish, while glue-qqp is applied to NLI tasks. In the second scenario, LoRAs are integrated for translating from German to French. Although there is no direct LoRA for German-to-French translation, the combined use of wmt16-deen for German-to-English and wmt14-enfr for English-to-French enables an effective German-to-French translation. The third scenario illustrates the fusion of distinct capabilities by combining Romanian translation with text generation: leveraging the wmt16-roen LoRA for Romanian comprehension and the common-gen LoRA for generating text, LoraRetriever successfully merges these diverse functionalities.
This demonstration emphasizes the framework's substantial ability to blend distinct LoRA capabilities, anticipating further exploration of capability fusion of LoRAs as a future direction.


\section{Conclusion}
This paper investigates a new problem of serving multiple LoRAs with a dynamically updated LoRA pool for downstream heterogeneous requests. To this end, we introduce a framework named LoraRetriver to identify and retrieve the appropriate LoRAs based on a specific input. Subsequently, we focus on the composition of these retrieved LoRAs to ensure a tailored and practical application in real-world situations. We also propose an efficient batch inference strategy to accommodate batched requests. Subsequent experiments have also demonstrated the effectiveness of our proposed LoraRetriever.

\section*{Limitation}
While promising, there are still some drawbacks of LoraRetriever. 
(1) User data privacy issues. When users upload LoRA, we need to use a small amount of training data (10-20 pieces) to represent the distribution of the LoRA model. In privacy-sensitive scenarios, representation with data may not be feasible. Aligning LoRA parameters and sample distributions in the embedding space in a manner that respects data privacy presents a worthwhile direction for future exploration.
(2) The proposed LoraRetriever framework is only suitable for multi-LoRA collaboration under the same model architecture. However, in reality, the model architecture chosen by the users themselves and the PEFT method are not necessarily the same, which is worth further research on how to design the corresponding collaborative mechanism for such scenarios.


\bibliography{anthology,custom}
\bibliographystyle{acl_natbib}

\appendix

\section{Details of Baseline Methods}
\label{sec:baseline}
(1) \textbf{Mixture of Experts} ~\cite{zhu2023sira,zadouri2023pushing,liu2023moelora,wang2022adamix,anonymous2024mole}. Many works have considered coordinating different adapters through MoE, and here we explored three distinct variants: one employing a soft mixture of experts and the other utilizing discrete routing (top1 and top3). Detailed information on the training aspects of MoE methods is provided in the Appendix.\ref{sec:moe_train}. 
(2) \textbf{SMEAR} ~\cite{muqeeth2023soft} introduces the concept of adaptive routing by performing a weighted average of different adapters' parameters to utilize various experts effectively. For the MoE and SMEAR baselines, challenges arise in scaling due to training confined to a limited set of LoRAs. Consequently, we strategically selected a dedicated LoRA expert for each domain to specialize in router training.
(3) \textbf{AdapterSoup} ~\cite{chronopoulou2023adaptersoup} uniformly selects the corresponding LoRAs for the entire downstream task, which lacks the ability to provide personalized service for diverse requests.
(4) \textbf{LoRAhub} ~\cite{huang2023lorahub} enables black-box optimization to learn the weights of various LoRA parameters, thereby facilitating weighted parameter averaging for specific downstream tasks. In our implementation, we conformed to the default setting, which entails randomly selecting 20 LoRAs from the available LoRA pool and performing weighted parameter averaging.
For the MoE, SMEAR, and LoRAhub approaches, we selected 20 data samples from the training datasets of all tasks to serve as their training data.

\section{Main Difference between LoraRetriever and Baseline Methods}
\begin{figure}[h]
    \centering
\includegraphics[width=.85\linewidth]{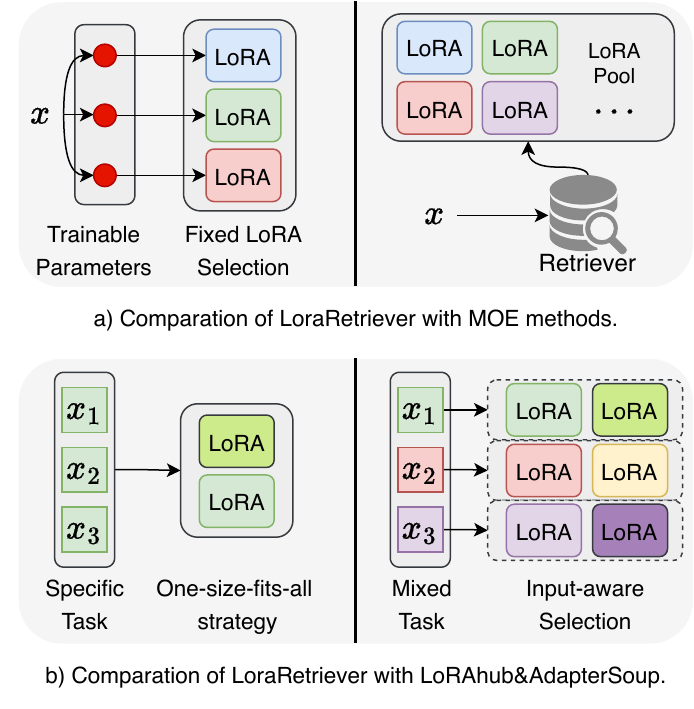}
    \caption{The advantages of LoraRetriever compared to previous methods. The above figure shows the difference between our method and MoE in expert routing; the figure below demonstrates the difference between our method's focus on mixed-task scenarios and previous methods that only targeted specific tasks.}
    \label{fig:compare}
\end{figure}

\section{Full Results}
\label{sec:full_results}
In Tab.\ref{tab:full_results}, we show the full results of the mixed-task scenario of all tasks.
\begin{table*}
\centering
\resizebox{.85\linewidth}{!}{  
\begin{tabular}{l c c c c c c c c c c c c c} 
\toprule
\multirow{2}{*}{Task / Llama-2-7b} &
\multirow{2}{*}{\begin{tabular}[c]{@{}c@{}}Perfect\\Selection\end{tabular}} &
\multicolumn{2}{c}{Selection} & 
\multicolumn{2}{c}{Fusion} & 
\multicolumn{2}{c}{Mixture} & 
\multirow{2}{*}{\begin{tabular}[c]{@{}c@{}}MoE\\Top1\end{tabular}} &
\multirow{2}{*}{\begin{tabular}[c]{@{}c@{}}MoE\\Top3\end{tabular}} &
\multirow{2}{*}{\begin{tabular}[c]{@{}c@{}}MoE\\Soft\end{tabular}} &
\multirow{2}{*}{\begin{tabular}[c]{@{}c@{}}SME-\\AR\end{tabular}} &
\multirow{2}{*}{\begin{tabular}[c]{@{}c@{}}Adapter\\Soup\end{tabular}} &
\multirow{2}{*}{\begin{tabular}[c]{@{}c@{}}LoRA\\Hub\end{tabular}}\\
\cline{3-8}
& & IID & OOD & IID & OOD & IID & OOD & & & & \\
\midrule
\multicolumn{12}{l}{\textbf{Struct to Text}} \\
WebNLG \textsuperscript{Rouge-1} & 71.2 & 67.0 & 53.9 & 49.4 & 45.4 & 57.8  & 53.9 & 45.1    & 47.6    & 49.1    & 51.1 & 3.9 & 32.5     \\
WebNLG \textsuperscript{Rouge-2} & 50.6 & 44.5 & 30.0 & 25.9 & 24.1 & 33.5  & 29.4 & 22.6    & 25.8    & 26.1    & 27.9 & 0.9 & 17.3     \\
WebNLG \textsuperscript{Rouge-l} & 64.4 & 60.9 & 49.1 & 45.5 & 41.0 & 52.3  & 49.6 & 40.0    & 41.9    & 43.3    & 45.4 & 3.9 & 31.1     \\
DART \textsuperscript{Rouge-1}  & 71.7 & 67.9 & 58.4 & 56.3 & 53.4 & 63.2  & 60.0 & 55.4    & 56.3    & 56.9    & 60.0 & 3.3 & 40.0     \\
DART \textsuperscript{Rouge-2}  & 49.1 & 45.8 & 34.9 & 32.3 & 30.6 & 36.6  & 35.4 & 30.3    & 31.0    & 30.8    & 33.0 & 1.3 & 20.1     \\
DART \textsuperscript{Rouge-l} & 64.6 & 61.1 & 52.4 & 50.3 & 47.9 & 56.3  & 52.4 & 49.7    & 50.8    & 50.2    & 54.8 & 3.3 & 35.2     \\
E2ENLG \textsuperscript{Rouge-1}    & 66.1 & 65.8 & 59.3 & 62.2 & 57.2 & 66.0  & 58.7 & 52.9    & 54.0    & 55.3    & 53.2 & 4.2 & 50.1     \\
E2ENLG \textsuperscript{Rouge-2}    & 40.0 & 39.4 & 34.1 & 34.7 & 32.0 & 38.8  & 32.1 & 26.9    & 27.6    & 28.8    & 27.5 & 2.4 & 26.3     \\
E2ENLG \textsuperscript{Rouge-l}   & 56.7 & 55.7 & 50.2 & 52.7 & 49.1 & 56.9  & 49.0 & 45.1    & 45.0    & 47.0    & 45.1 & 4.2 & 42.2     \\
CommonGen \textsuperscript{Rouge-1} & 46.9 & 44.7 & 29.0 & 29.9 & 27.7 & 36.5  & 29.0 & 29.0    & 29.3    & 30.1    & 27.6 & 6.6 & 19.8     \\
CommonGen \textsuperscript{Rouge-2} & 18.8 & 18.3 & 7.3 & 9.9  & 7.2 & 11.1  & 8.6  & 7.7     & 7.1     & 9.3     & 8.4 & 0.0 & 6.9 \\
CommonGen \textsuperscript{Rouge-l} & 42.5 & 40.5 & 24.0 & 25.8 & 23.3 & 32.7  & 24.8 & 24.4    & 25.1    & 26.3    & 24.3 & 6.6 & 18.0     \\
\midrule
\multicolumn{12}{l}{\textbf{Translation}} \\
Paracrawl-enes   & 24.3 & 24.2 & 20.3 & 22.9 & 22.3 & 22.8  & 22.1 & 18.0    & 18.8    & 19.5    & 21.6 & 4.5 & 16.4     \\
WMT'16-tren   & 3.2  & 3.1  & 2.6 & 3.5  & 3.3 & 3.7   & 2.6  & 3.5     & 3.2     & 3.4     & 3.2 & 0.0 & 2.0 \\
WMT'16-ruen   & 10.8 & 10.4 & 9.8 & 9.2  & 9.3 & 11.0  & 10.8 & 6.2     & 7.8     & 8.3     & 7.3 & 0.0 & 4.8 \\
WMT'16-deen   & 18.9 & 18.7 & 20.3 & 17.9 & 18.8 & 18.8  & 18.7 & 11.6    & 14.0    & 14.7    & 16.6 & 1.1 & 11.4     \\
WMT'16-fien   & 6.5  & 6.5  & 7.0 & 7.2  & 7.1 & 7.3   & 7.8  & 6.2     & 6.2     & 6.1     & 6.5 & 0.7 & 4.3 \\
WMT'16-roen   & 13.9 & 14.0 & 12.3 & 12.8 & 13.3 & 13.1  & 12.2 & 9.8     & 10.7    & 10.1    & 10.3 & 0.3 & 8.0 \\
WMT'14-enfr  & 16.5 & 16.1 & 16.9 & 17.7 & 18.0 & 17.8  & 18.0 & 15.9    & 17.3    & 17.1    & 16.4 & 3.5 & 15.2     \\
WMT'16-csen   & 10.7 & 9.4  & 7.0 & 6.1  & 6.2 & 8.3   & 5.8  & 4.7     & 6.3     & 6.3     & 6.3 & 0.8 & 6.1 \\
\midrule
\multicolumn{12}{l}{ \textbf{COMMONSENSE}}       \\
StoryCloze & 72.0 & 62.0 & 42.0 & 72.0 & 68.0 & 84.0  & 58.0 & 74.0    & 70.0    & 70.0    & 68.0 & 62.0     & 48.0     \\
PIQA  & 46.0 & 46.0 & 32.0 & 34.0 & 36.0 & 38.0  & 34.0 & 40.0    & 38.0    & 38.0    & 36.0 & 38.0     & 0.0 \\
COPA  & 86.0 & 74.0 & 68.0 & 78.0 & 70.0 & 80.0  & 68.0 & 72.0    & 70.0    & 72.0    & 70.0 & 56.0     & 22.0     \\
HellaSwag   & 46.0 & 40.0 & 42.0 & 20.0 & 18.0 & 44.0  & 40.0 & 32.0    & 30.0    & 26.0    & 26.0 & 28.0     & 0.0 \\
\hline
\multicolumn{12}{l}{\textbf{sentiment}}    \\
SST-2  & 98.0 & 98.0 & 96.0 & 74.0 & 78.0 & 96.0  & 94.0 & 56.0    & 68.0    & 66.0    & 66.0 & 74.0     & 0.0 \\
Yelp  & 98.0 & 94.0 & 94.0 & 96.0 & 96.0 & 98.0  & 98.0 & 86.0    & 90.0    & 86.0    & 84.0 & 80.0     & 0.0 \\
IMDB & 96.0 & 96.0 & 96.0 & 92.0 & 82.0 & 96.0  & 96.0 & 76.0    & 80.0    & 80.0    & 84.0 & 80.0     & 0.0 \\
sentiment140 & 68.0 & 70.0 & 70.0 & 54.0 & 58.0 & 68.0  & 74.0 & 62.0    & 62.0    & 66.0    & 62.0 & 60.0     & 2.0 \\
\hline
\multicolumn{12}{l}{\textbf{READING Comp.}}     \\
MultiRC     & 68.0 & 52.0 & 38.0 & 44.0 & 44.0 & 48.0  & 44.0 & 54.0    & 52.0    & 50.0    & 48.0 & 40.0     & 6.0 \\
SQuADv2    & 62.0 & 56.0 & 12.0 & 30.0 & 20.0 & 22.0  & 16.0 & 24.0    & 24.0    & 26.0    & 22.0 & 16.0     & 0.0 \\
SQuADv1    & 68.0 & 66.0 & 68.0 & 64.0 & 64.0 & 62.0  & 68.0 & 68.0    & 70.0    & 66.0    & 66.0 & 54.0     & 4.0 \\
OBQA  & 82.0 & 68.0 & 58.0 & 64.0 & 60.0 & 78.0  & 66.0 & 62.0    & 64.0    & 66.0    & 60.0 & 40.0     & 0.0 \\
BoolQ & 84.0 & 60.0 & 60.0 & 68.0 & 70.0 & 80.0  & 76.0 & 74.0    & 68.0    & 76.0    & 70.0 & 72.0     & 6.0 \\
drop  & 40.0 & 8.0  & 6.0 & 14.0 & 12.0 & 18.0  & 14.0 & 10.0    & 8.0     & 8.0     & 8.0 & 22.0     & 0.0 \\
\hline
\multicolumn{12}{l}{\textbf{CLOSE-BOOK QA}}           \\
NQ & 18.0 & 16.0 & 10.0 & 16.0 & 14.0 & 16.0  & 10.0 & 12.0    & 12.0    & 12.0    & 4.0 & 12.0     & 0.0 \\
ARC-e   & 50.0 & 56.0 & 70.0 & 54.0 & 56.0 & 66.0  & 82.0 & 58.0    & 58.0    & 60.0    & 58.0 & 48.0     & 0.0 \\
ARC-c     & 46.0 & 42.0 & 46.0 & 34.0 & 34.0 & 50.0  & 46.0 & 46.0    & 42.0    & 42.0    & 42.0 & 24.0     & 0.0 \\
TriviaQa   & 66.0 & 46.0 & 46.0 & 60.0 & 46.0 & 48.0  & 56.0 & 46.0    & 42.0    & 46.0    & 24.0 & 42.0     & 4.0 \\
\hline
\multicolumn{12}{l}{\textbf{COREFERENCE}}     \\
DPR & 54.0 & 50.0 & 50.0 & 56.0 & 60.0 & 68.0  & 56.0 & 64.0    & 60.0    & 62.0    & 62.0 & 46.0     & 2.0 \\
WSC   & 50.0 & 50.0 & 42.0 & 38.0 & 46.0 & 58.0  & 42.0 & 58.0    & 58.0    & 52.0    & 54.0 & 40.0     & 0.0 \\
\hline
\multicolumn{12}{l}{\textbf{READ. COOMP. W/ COMMONSENSE}}  \\
CosmosQa   & 68.0 & 68.0 & 34.0 & 46.0 & 32.0 & 50.0  & 46.0 & 44.0    & 46.0    & 44.0    & 38.0 & 14.0     & 6.0 \\
record & 70.0 & 70.0 & 26.0 & 24.0 & 6.0 & 42.0  & 34.0 & 18.0    & 12.0    & 14.0    & 8.0 & 14.0     & 0.0 \\
\hline
\multicolumn{12}{l}{\textbf{PARAPHRASE}}    \\
Paws Wiki   & 90.0 & 64.0 & 40.0 & 44.0 & 42.0 & 56.0  & 46.0 & 56.0    & 50.0    & 48.0    & 54.0 & 60.0     & 2.0 \\
QQP    & 74.0 & 74.0 & 68.0 & 66.0 & 60.0 & 80.0  & 58.0 & 50.0    & 40.0    & 36.0    & 28.0 & 54.0     & 0.0 \\
MRPC   & 60.0 & 58.0 & 58.0 & 60.0 & 62.0 & 60.0  & 58.0 & 42.0    & 44.0    & 40.0    & 42.0 & 60.0     & 2.0 \\
STSB  & 38.0 & 36.0 & 16.0 & 12.0 & 12.0 & 30.0  & 20.0 & 20.0    & 20.0    & 20.0    & 14.0 & 12.0     & 0.0 \\
\hline
\multicolumn{12}{l}{\textbf{NLI}}     \\
CB    & 88.9 & 80.0 & 62.2 & 77.8 & 57.8 & 86.7  & 66.7 & 68.9    & 64.4    & 68.9    & 62.2 & 55.6     & 13.3     \\
WNLI  & 70.0 & 68.0 & 46.0 & 44.0 & 50.0 & 60.0  & 54.0 & 56.0    & 56.0    & 42.0    & 44.0 & 52.0     & 0.0 \\
ANLI-r1     & 50.0 & 50.0 & 50.0 & 40.0 & 42.0 & 40.0  & 42.0 & 40.0    & 40.0    & 36.0    & 38.0 & 38.0     & 24.0     \\
ANLI-r2     & 46.0 & 46.0 & 46.0 & 32.0 & 36.0 & 46.0  & 46.0 & 40.0    & 36.0    & 38.0    & 32.0 & 46.0     & 20.0     \\
ANLI-r3     & 46.0 & 42.0 & 38.0 & 38.0 & 40.0 & 44.0  & 50.0 & 28.0    & 32.0    & 34.0    & 38.0 & 40.0     & 24.0     \\
MNLI-m & 88.0 & 84.0 & 88.0 & 62.0 & 66.0 & 80.0  & 88.0 & 48.0    & 54.0    & 50.0    & 56.0 & 76.0     & 0.0 \\
MNLI-mm   & 92.0 & 90.0 & 94.0 & 64.0 & 82.0 & 88.0  & 90.0 & 48.0    & 48.0    & 50.0    & 60.0 & 84.0     & 2.0 \\
SNLI  & 96.0 & 84.0 & 84.0 & 56.0 & 58.0 & 90.0  & 92.0 & 54.0    & 52.0    & 54.0    & 54.0 & 82.0     & 0.0 \\
QNLI  & 94.0 & 94.0 & 26.0 & 46.0 & 48.0 & 74.0  & 38.0 & 56.0    & 56.0    & 54.0    & 60.0 & 70.0     & 0.0 \\
RTE   & 52.0 & 62.0 & 72.0 & 54.0 & 58.0 & 70.0  & 76.0 & 64.0    & 58.0    & 56.0    & 64.0 & 80.0     & 22.0     \\
\bottomrule
\end{tabular}
}
\caption{Mixed Tasks evaluation on both NLU \& NLG tasks. ``OOD" indicates that during retrieval, we masked the corresponding task's LoRA for testing generalization when facing unknown tasks.}
\label{tab:full_results}
\end{table*}

\begin{table*}
\centering
\resizebox{.85\linewidth}{!}{  
\begin{tabular}{l c c c c c c c c c c c c c} 
\toprule
\multirow{2}{*}{Task / Llama-2-13b} &
\multirow{2}{*}{\begin{tabular}[c]{@{}c@{}}Perfect\\Selection\end{tabular}} &
\multicolumn{2}{c}{Selection} & 
\multicolumn{2}{c}{Fusion} & 
\multicolumn{2}{c}{Mixture} & 
\multirow{2}{*}{\begin{tabular}[c]{@{}c@{}}MoE\\Top1\end{tabular}} &
\multirow{2}{*}{\begin{tabular}[c]{@{}c@{}}MoE\\Top3\end{tabular}} &
\multirow{2}{*}{\begin{tabular}[c]{@{}c@{}}MoE\\Soft\end{tabular}} &
\multirow{2}{*}{\begin{tabular}[c]{@{}c@{}}SME-\\AR\end{tabular}} &
\multirow{2}{*}{\begin{tabular}[c]{@{}c@{}}Adapter\\Soup\end{tabular}} &
\multirow{2}{*}{\begin{tabular}[c]{@{}c@{}}LoRA\\Hub\end{tabular}}\\
\cline{3-8}
& & IID & OOD & IID & OOD & IID & OOD & & & & \\
\midrule
\multicolumn{12}{l}{\textbf{Struct to Text}} \\
WebNLG \textsuperscript{Rouge-1} & 72.6 & 68.5 & 51.9      & 55.2 & 51.3      & 59.7  & 53.7 & 47.0    & 47.8    & 48.3    & 49.7      & 6.6      & 34.2          \\
WebNLG \textsuperscript{Rouge-2} & 51.4 & 47.5 & 28.6      & 30.1 & 27.4      & 35.6  & 29.1 & 25.5    & 26.4    & 27.0    & 26.3      & 3.2      & 16.8          \\
WebNLG \textsuperscript{Rouge-l} & 66.0 & 62.4 & 48.3      & 49.4 & 46.2      & 55.1  & 49.0 & 42.5    & 44.3    & 43.5    & 44.3      & 6.3      & 32.4          \\
DART \textsuperscript{Rouge-1}  & 74.0 & 67.0 & 57.0      & 60.4 & 58.7      & 62.6  & 60.6 & 57.9    & 57.0    & 58.7    & 58.9      & 12.5     & 43.3          \\
DART \textsuperscript{Rouge-2}  & 54.6 & 45.9 & 33.6      & 37.4 & 35.2      & 38.9  & 37.3 & 32.6    & 33.2    & 34.1    & 34.3      & 5.9      & 25.5          \\
DART \textsuperscript{Rouge-l} & 67.7 & 61.2 & 50.0      & 54.0 & 52.2      & 55.0  & 53.4 & 50.9    & 50.9    & 51.8    & 51.5      & 11.8     & 38.1          \\
E2ENLG \textsuperscript{Rouge-1}  & 66.4 & 66.1 & 59.2      & 65.6 & 61.7      & 66.7  & 63.9 & 52.8    & 53.9    & 55.5    & 55.1      & 2.5      & 58.6          \\
E2ENLG \textsuperscript{Rouge-2}   & 39.6 & 39.3 & 32.8      & 36.6 & 33.5      & 39.0  & 36.4 & 27.8    & 28.4    & 28.3    & 28.8      & 0.9      & 31.6          \\
E2ENLG \textsuperscript{Rouge-l}  & 56.7 & 56.4 & 48.9      & 53.4 & 49.7      & 56.2  & 53.7 & 43.0    & 44.5    & 45.1    & 45.4      & 2.2      & 48.2          \\
CommonGen \textsuperscript{Rouge-1} & 48.5 & 48.9 & 29.3      & 29.5 & 27.0      & 41.7  & 30.0 & 29.5    & 29.3    & 31.5    & 29.5      & 6.9      & 21.2          \\
CommonGen \textsuperscript{Rouge-2} & 17.7 & 20.3 & 8.2 & 12.6 & 11.3      & 17.0  & 9.5  & 12.2    & 12.3    & 13.5    & 11.5      & 0.0      & 8.8 \\
CommonGen \textsuperscript{Rouge-l} & 44.3 & 44.1 & 24.4      & 26.7 & 24.6      & 36.7  & 25.3 & 28.0    & 27.9    & 30.3    & 27.8      & 5.4      & 19.7          \\
\midrule
\multicolumn{12}{l}{\textbf{Translation}} \\
Paracrawl-enes   & 24.4 & 25.4 & 23.1      & 28.3 & 27.2      & 27.4  & 25.8 & 21.8    & 21.5    & 20.0    & 24.2      & 4.0      & 19.0          \\
WMT'16-tren    & 2.9  & 2.4  & 1.2 & 3.7  & 3.2 & 3.4   & 2.7  & 3.0     & 3.3     & 2.7     & 3.3 & 0.0      & 1.9 \\
WMT'16-ruen   & 11.8 & 11.5 & 10.3      & 11.8 & 11.5      & 10.5  & 12.0 & 8.8     & 9.9     & 9.3     & 8.8 & 0.0      & 8.1 \\
WMT'16-deen   & 19.9 & 19.9 & 20.7      & 20.2 & 20.7      & 20.2  & 20.2 & 16.5    & 18.6    & 18.1    & 18.3      & 2.2      & 15.1          \\
WMT'16-fien   & 7.3  & 6.8  & 5.1 & 9.8  & 7.3 & 8.7   & 7.8  & 8.0     & 8.3     & 8.4     & 8.3 & 0.0      & 6.3 \\
WMT'16-roen   & 14.0 & 13.9 & 10.9      & 11.6 & 11.1      & 17.4  & 13.3 & 7.9     & 9.1     & 9.3     & 9.3 & 0.0      & 7.4 \\
WMT'14-enfr  & 16.6 & 17.1 & 18.4      & 20.7 & 21.1      & 18.7  & 20.9 & 17.4    & 17.5    & 18.7    & 17.2      & 0.4      & 15.1          \\
WMT'16-csen   & 6.6  & 6.2  & 11.6      & 11.1 & 10.5      & 10.3  & 10.1 & 10.6    & 10.7    & 9.0     & 9.4 & 0.0      & 8.8 \\
\midrule
\multicolumn{12}{l}{ \textbf{COMMONSENSE}}       \\
StoryCloze & 96.0 & 80.0 & 56.0      & 90.0 & 76.0      & 80.0  & 76.0 & 96.0    & 98.0    & 94.0    & 92.0      & 18.0     & 64.0          \\
PIQA  & 48.0 & 52.0 & 30.0      & 46.0 & 46.0      & 46.0  & 46.0 & 42.0    & 40.0    & 36.0    & 38.0      & 14.0     & 10.0          \\
COPA  & 76.0 & 74.0 & 68.0      & 74.0 & 74.0      & 78.0  & 76.0 & 72.0    & 80.0    & 80.0    & 76.0      & 22.0     & 60.0          \\
HellaSwag   & 58.0 & 30.0 & 36.0      & 34.0 & 28.0      & 52.0  & 44.0 & 50.0    & 46.0    & 46.0    & 38.0      & 16.0     & 2.0 \\
\hline
\multicolumn{12}{l}{\textbf{sentiment}}    \\
SST-2  & 98.0 & 98.0 & 98.0      & 86.0 & 86.0      & 98.0  & 100.0      & 98.0    & 96.0    & 98.0    & 96.0      & 82.0     & 26.0          \\
Yelp  & 98.0 & 98.0 & 100.0     & 94.0 & 92.0      & 98.0  & 98.0 & 98.0    & 98.0    & 98.0    & 98.0      & 82.0     & 0.0 \\
IMDB & 96.0 & 98.0 & 98.0      & 88.0 & 82.0      & 98.0  & 98.0 & 100.0   & 100.0   & 100.0   & 98.0      & 90.0     & 4.0 \\
sentiment140 & 68.0 & 68.0 & 68.0      & 80.0 & 74.0      & 72.0  & 70.0 & 64.0    & 64.0    & 64.0    & 64.0      & 64.0     & 14.0          \\
\hline
\multicolumn{12}{l}{\textbf{READING Comp.}}     \\
MultiRC   & 88.0 & 72.0 & 36.0      & 64.0 & 42.0      & 66.0  & 44.0 & 44.0    & 48.0    & 40.0    & 40.0      & 72.0     & 2.0 \\
SQuADv1    & 74.0 & 62.0 & 62.0      & 58.0 & 58.0      & 60.0  & 60.0 & 64.0    & 66.0    & 64.0    & 60.0      & 42.0     & 8.0 \\
SQuADv2    & 66.0 & 58.0 & 34.0      & 38.0 & 24.0      & 34.0  & 24.0 & 26.0    & 22.0    & 24.0    & 24.0      & 36.0     & 0.0 \\
OBQA  & 86.0 & 80.0 & 72.0      & 76.0 & 76.0      & 88.0  & 82.0 & 80.0    & 76.0    & 78.0    & 76.0      & 48.0     & 0.0 \\
BoolQ & 84.0 & 68.0 & 66.0      & 84.0 & 82.0      & 78.0  & 78.0 & 86.0    & 84.0    & 88.0    & 86.0      & 74.0     & 10.0          \\
Drpo & 58.0 & 22.0 & 18.0      & 20.0 & 14.0      & 36.0  & 20.0 & 22.0    & 24.0    & 20.0    & 22.0      & 20.0     & 0.0 \\
\hline
\multicolumn{12}{l}{\textbf{CLOSE-BOOK QA}}           \\
NQ & 30.0 & 28.0 & 12.0      & 22.0 & 14.0      & 24.0  & 16.0 & 12.0    & 10.0    & 12.0    & 10.0      & 6.0      & 2.0 \\
ARC-e   & 90.0 & 90.0 & 88.0      & 92.0 & 92.0      & 94.0  & 94.0 & 90.0    & 90.0    & 86.0    & 90.0      & 58.0     & 8.0 \\
ARC-c     & 68.0 & 66.0 & 58.0      & 64.0 & 60.0      & 68.0  & 70.0 & 68.0    & 60.0    & 68.0    & 62.0      & 38.0     & 4.0 \\
TriviaQa   & 68.0 & 56.0 & 54.0      & 70.0 & 66.0      & 66.0  & 64.0 & 68.0    & 70.0    & 68.0    & 68.0      & 36.0     & 12.0          \\
\hline
\multicolumn{12}{l}{\textbf{COREFERENCE}}     \\
DPR & 90.0 & 90.0 & 72.0      & 68.0 & 66.0      & 88.0  & 72.0 & 76.0    & 76.0    & 68.0    & 72.0      & 52.0     & 20.0          \\
WSC  & 58.0 & 60.0 & 58.0      & 42.0 & 52.0      & 64.0  & 56.0 & 46.0    & 48.0    & 44.0    & 42.0      & 58.0     & 0.0 \\
\hline
\multicolumn{12}{l}{\textbf{READ. COOMP. W/ COMMONSENSE}}  \\
CosmosQa  & 84.0 & 82.0 & 38.0      & 80.0 & 76.0      & 82.0  & 70.0 & 74.0    & 74.0    & 74.0    & 80.0      & 8.0      & 28.0          \\
record & 80.0 & 78.0 & 28.0      & 34.0 & 22.0      & 74.0  & 46.0 & 28.0    & 22.0    & 24.0    & 18.0      & 18.0     & 0.0 \\
\hline
\multicolumn{12}{l}{\textbf{PARAPHRASE}}    \\
Paws Wiki  & 92.0 & 70.0 & 52.0      & 74.0 & 50.0      & 88.0  & 74.0 & 54.0    & 66.0    & 54.0    & 62.0      & 90.0     & 8.0 \\
QQP   & 90.0 & 88.0 & 76.0      & 66.0 & 56.0      & 84.0  & 62.0 & 70.0    & 66.0    & 60.0    & 66.0      & 78.0     & 2.0 \\
MRPC  & 84.0 & 70.0 & 62.0      & 64.0 & 64.0      & 78.0  & 64.0 & 62.0    & 62.0    & 62.0    & 64.0      & 82.0     & 0.0 \\
STSB  & 44.0 & 44.0 & 20.0      & 18.0 & 12.0      & 34.0  & 22.0 & 14.0    & 16.0    & 14.0    & 16.0      & 6.0      & 0.0 \\
\hline
\multicolumn{12}{l}{\textbf{NLI}}     \\
CB    & 100.0      & 91.1 & 84.4      & 84.4 & 80.0      & 93.3  & 88.9 & 73.3    & 82.2    & 75.6    & 77.8      & 73.3     & 13.3          \\
WNLI  & 72.0 & 72.0 & 62.0      & 70.0 & 68.0      & 76.0  & 66.0 & 58.0    & 66.0    & 56.0    & 56.0      & 76.0     & 6.0 \\
ANLI-r1    & 70.0 & 70.0 & 70.0      & 56.0 & 50.0      & 64.0  & 68.0 & 48.0    & 44.0    & 44.0    & 46.0      & 50.0     & 12.0  \\
ANLI-r2    & 64.0 & 56.0 & 56.0      & 38.0 & 36.0      & 60.0  & 60.0 & 48.0    & 48.0    & 48.0    & 42.0      & 48.0     & 8.0 \\
ANLI-r3     & 68.0 & 56.0 & 56.0      & 46.0 & 46.0      & 62.0  & 60.0 & 46.0    & 48.0    & 56.0    & 58.0      & 48.0     & 20.0   \\
MNLI-m & 88.0 & 90.0 & 88.0      & 88.0 & 88.0      & 86.0  & 88.0 & 90.0    & 94.0    & 88.0    & 80.0      & 96.0     & 4.0 \\
MNLI-mm  & 90.0 & 90.0 & 94.0      & 94.0 & 92.0      & 94.0  & 100.0      & 94.0    & 98.0    & 88.0    & 88.0      & 80.0     & 2.0 \\
SNLI  & 90.0 & 88.0 & 88.0      & 76.0 & 74.0      & 90.0  & 92.0 & 96.0    & 96.0    & 94.0    & 86.0      & 80.0     & 16.0          \\
QNLI  & 94.0 & 94.0 & 30.0      & 68.0 & 56.0      & 74.0  & 58.0 & 60.0    & 62.0    & 60.0    & 56.0      & 56.0     & 32.0          \\
RTE   & 88.0 & 82.0 & 74.0      & 78.0 & 74.0      & 82.0  & 76.0 & 64.0    & 72.0    & 64.0    & 76.0      & 68.0     & 36.0          \\
\bottomrule
\end{tabular}
}
\caption{Mixed Tasks evaluation on both NLU \& NLG tasks. ``OOD" indicates that during retrieval, we masked the corresponding task's LoRA for testing generalization when facing unknown tasks.}
\label{tab:full_results_13b}
\end{table*}

\section{PyTorch-style pseudocode for batch inference}
\label{sec:pseudocode}
The batch inference process can be easily achieved through a few lines of einsum operation. We show the PyTorch style pseudocode in Alg.\ref{alg:batch_inference}.
\begin{algorithm}
    \PyComment{X: (b,l,d), M: (b,p)} \\
    \PyComment{A: (p,r,d), B: (p,d,r)} \\
    \PyComment{LoRA fusion computation} \\
    \PyCode{FA=torch.einsum('bp,prd->brd',M,A)} \\
    \PyCode{FB=torch.einsum('bp,pdr->bdr',M,B)} \\
    \PyCode{mid=torch.einsum('bld,brd->blr',X,FA)} \\
    \PyCode{res=torch.einsum('blr,bdr->bld',mid,FB)} \\
    \PyComment{LoRA mixture computation} \\
    \PyCode{mid=torch.einsum('bld,prd->blpr',X,A)} \\
    \PyCode{mid=torch.einsum('blpr,pdr->blpd',mid,B)} \\
    \PyCode{res=torch.einsum('blpd,bp->bld',mid,M)} \\
\caption{PyTorch-style pseudocode for batch inference}
\label{alg:batch_inference}
\end{algorithm}

\begin{figure*}
    \centering
    \includegraphics[width=.8\linewidth]{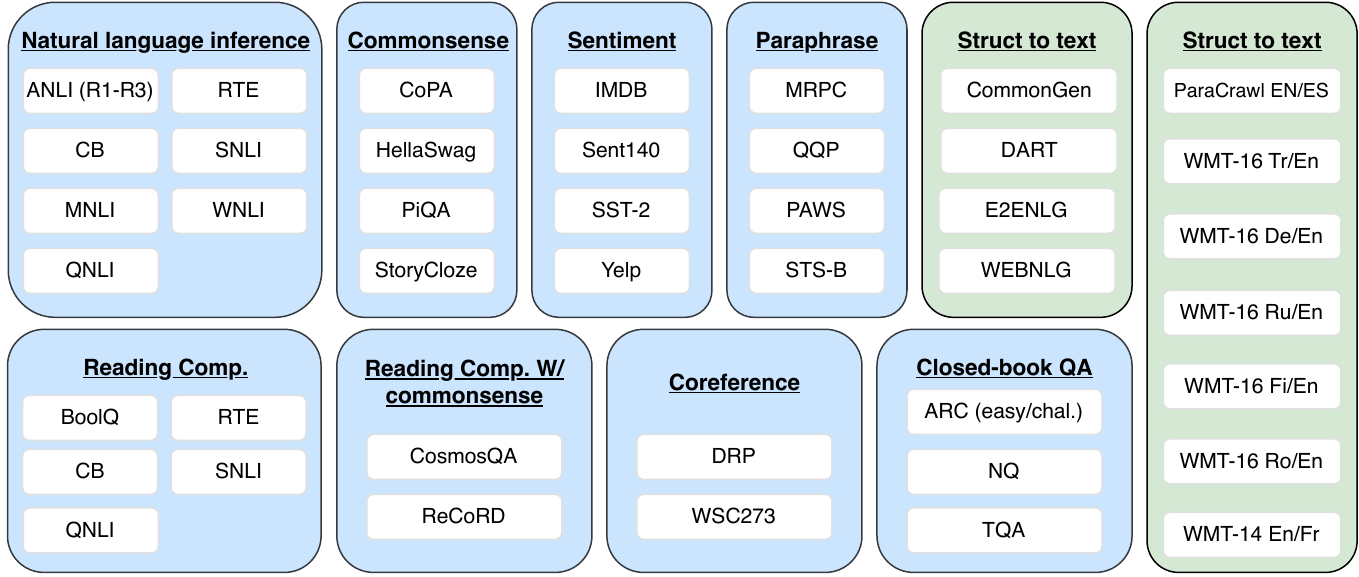}
    \caption{Datasets and task clusters used to train LoRAs and generate mixed-task evaluation set in this paper (NLU tasks in blue; NLG tasks in green).
    }
    \label{fig:tasks}
\end{figure*}
\section{Details of Training and Evaluation datasets}
\label{sec:eval_dataset}
We leverage a subset of flan-v2 datasets ~\cite{wei2021finetuned} as shown in Fig.\ref{fig:tasks} for LoRA expert training and mixed-task dataset generation. We summarize the details of the used datasets as follows:

\textbf{Struct-to-Text Conversion}: This task evaluates the capability to generate natural language descriptions from structured data inputs. We use the following datasets: (1) CommonGen; (2) DART; (3) E2ENLG; (4) WebNLG;

\textbf{Translation}: Translation involves converting text from one language to another, maintaining the original meaning and nuances. We use the following datasets: (1) En-Fr from WMT'14; En-De, En-Tr, En-Ru, En-Fi, En-Ro from WMT'16; (3) En-Es from Paracrawl.

\textbf{Commonsense Reasoning}: This involves assessing the ability to apply physical or scientific principles alongside common sense in reasoning tasks. We use the following datasets: (1) COPA, (2) HellaSwag, (3) PiQA, and (4) StoryCloze.

\textbf{Sentiment Analysis}: A fundamental task in natural language processing (NLP) that determines the sentiment polarity (positive or negative) of a given text. We use the following datasets: (1) IMDB, (2) Sentiment140, (3) SST-2, and (4) Yelp.

\textbf{Closed-Book Question Answering}: This task challenges models to answer questions about general knowledge without direct access to external information sources. We use the following datasets: (1) ARC, (2) NQ, and (3) TriviaQA.

\textbf{Paraphrase Detection}: This task requires models to ascertain whether two sentences convey the same meaning, indicating semantic equivalence. We use the following datasets: (1) MRPC, (2) QQP, and (3) Paws Wiki.

\textbf{Coreference Resolution}: Involves identifying instances within a text that refer to the same entity, demonstrating an understanding of textual context. We use the following datasets: (1) DPR and (2) WSC273.

\textbf{Reading comprehension}: Assesses the capability to derive answers to questions from a provided text containing relevant information. We use the following datasets: (1) BoolQ, (2) DROP, (3) MultiRC, (4) OBQA, (5) SQuADv1, (6) SQuADv2.

\textbf{Reading Comprehension with Commonsense}: Merges traditional reading comprehension skills with commonsense reasoning, requiring understanding beyond the explicit text. We use the following datasets: (1) CosmosQA; (2) ReCoRD.

\textbf{Natural Language Inference}: Focuses on deducing the relationship between two sentences, determining if the second sentence logically follows from, contradicts, or is unrelated to the first sentence. We use the following datasets: (1) ANLI, (2) CB; (3) MNLI; (4) QNLI; (5) SNLI; (6) WNLI; (7) RTE.

\section{Implementation Details of Baseline Methods}
\label{sec:moe_train}
\subsection{MoE baselines}
We use $E$ to denote the LoRA expert and $R$ to denote the router. The MoE methods can be expressed in the following way:
\begin{equation}
    y = \sum_{i=i}^k R(x)_i E_i(x).
\end{equation}
We implied two variants of the MoE routing mechanism. (1) \textbf{Dense Gating.} Following ~\cite{zadouri2023pushing}, the router network consists of a dense layer with trainable parameter $W_g$, and the gating score could be obtained through a softmax function by:
\begin{equation}
    s_i = R(x)_i = softmax(W_g^T x),
\end{equation}

(2)\textbf{Sparse Gate}. To maintain the sparsity while training, we leverage the Gumbel softmax trick as ~\cite{muqeeth2023soft, nie2021dense}, where the router can be written as:
\begin{equation}
    \hat{R}(x)_i = \frac{(log(R(x)_i)+g_i)/ \tau}{\sum_{i=1}^k exp((log(R(x)_i)+g_i)/\tau)}
\end{equation}
where $g_i \sim \text{Gumbel}(0,1)$ and $\tau$ is the temperature. 

Due to MoE not being easily scalable and arbitrarily adding new LORAs, we randomly selected a LoRA as an expert for each task cluster in the experiment and trained the corresponding Router's parameters. We randomly selected 20 samples for each task during training to form a unified dataset for parameter training.

\subsection{SMEAR}
SMEAR ~\cite{muqeeth2023soft} does not perform routing aggregation on the Adapter output but rather aggregates the Adapter at the parameter level. We adopt the same setting as the MoE methods, and the results could be calculated in the following way:
\begin{equation}
    \Theta_{SMEAR} = \sum_{i=i}^k R(x)_i \Theta_i,
\end{equation}
where $\Theta_i$ denote the parameter of the LoRA-$i$. 

\subsection{AdapterSoup}
AdapterSoup ~\cite{chronopoulou2023adaptersoup}, for new downstream tasks, retrieves the parameters that need to be involved in aggregation through sentence bert and performs weight-space averaging on these parameters to adapt to the new domain. We have uniformly retrieved 3 LoRAs for mixed-task to test their capabilities under mixed-task conditions.

\subsection{LoRAhub}
LoRAhub ~\cite{huang2023lorahub} also aggregates 20 LoRAs randomly for new downstream tasks. In order to learn the weight of LoRA, a black-box optimization method is employed to learn the weight of each LoRA without calculating the gradients of the large model. It performs weighted averaging at the parameter level. Similar to the training process of MoE, we randomly selected 20 samples for each task to form a unified training dataset for black-box optimization.

\section{More Related Works}
\textbf{Personalized LoRA serving.}
\citet{sheng2023s} propose S-LoRA to discuss serving thousands of concurrent LoRA. The framework targets scenarios in which multiple tasks must be handled simultaneously without compromising the efficiency of the base models. \citet{wen2023batched} propose FLoRA, which enables efficient batching of diverse request types in low-rank adaptation (LoRA) of foundation models. These studies discuss how to deploy or train personalized LoRAs. However, these methods can only utilize a single user-specified LoRA during inference, failing to fully leverage the combination of LoRAs from different tasks. Moreover, the primary focus of these discussions is on computational strategies in GPUs and training strategies, which are orthogonal to the routing strategies with which we are concerned.


\end{document}